\documentclass{article}

\usepackage{spconf}

\usepackage{graphicx}
\graphicspath{{fig/}}

\usepackage{booktabs}
\usepackage{tabularx}
\usepackage{multirow}
\newcommand{\mytable}{
    \centering
    \renewcommand{\arraystretch}{1.2}
    }
\newcolumntype{C}{>{\centering\arraybackslash}X}
\newcolumntype{L}{>{\raggedright\arraybackslash}X}
\newcolumntype{R}{>{\raggedleft\arraybackslash}X}
\newcolumntype{P}[1]{>{\raggedright\arraybackslash}p{#1}}

\usepackage{amsmath}
\usepackage{amssymb}
\renewcommand{\vec}[1]{\boldsymbol{{#1}}}

\newcommand{\pten}{$P@\textrm{10}$ }

\usepackage{microtype}
\usepackage{url}
\usepackage{xcolor}
\newcommand{\system}[1]{{\small \textsc{#1}}}
\newcommand\blfootnote[1]{\begingroup
                          \renewcommand\thefootnote{}\footnote{#1}
                          \addtocounter{footnote}{-1}
                          \endgroup}

\usepackage{morefloats}

\usepackage{cite}
\title{Semantic query-by-example speech search using visual grounding}
\name{Herman Kamper$^1$ \qquad Aristotelis Anastassiou$^1$ \qquad Karen Livescu$^2$} 
\address{$^1$E\&E Engineering, Stellenbosch University, South Africa \& $^2$TTI-Chicago, USA \\
         {\small \tt kamperh@sun.ac.za, aristaki@me.com, klivescu@ttic.edu}}

\usepackage[prependcaption,textsize=scriptsize]{todonotes}
\begin{document}

\maketitle

\begin{abstract}
A number of recent studies have started to investigate how speech systems can be trained on untranscribed speech by leveraging accompanying images at training time. Examples of tasks include keyword prediction and within- and across-mode retrieval. Here we consider how such models can be used for query-by-example (QbE) search, the task of retrieving utterances relevant to a given spoken query. We are particularly interested in {\it semantic} QbE, where the task is not only to retrieve utterances containing exact instances of the query, but also utterances whose meaning is relevant to the query. We follow a segmental QbE approach where variable-duration speech segments (queries, search utterances) are mapped to fixed-dimensional embedding vectors. We show that a QbE system using an embedding function trained on visually grounded speech data outperforms a purely acoustic QbE system in terms of both exact and semantic retrieval performance.
\end{abstract}

\section{Introduction}
\label{sec:intro}


While the field of speech processing has made great strides
for tasks and domains with large amounts of available training data, lower-data domains and languages are still not adequately addressed.  This
has led many to explore alternative, weaker sources of supervision when labelled data is not available~\cite{synnaeve+etal_slt14,palaz+etal_interspeech16,bansal+etal_interspeech18}.  One form of weak supervision that has seen recent success is visual grounding: the use of images paired with speech data~\cite{synnaeve+etal_nipsworkshop14,harwath+etal_nips16,chrupala+etal_acl17,scharenborg+etal_icassp18,kamper+roth_sltu18}.  While we do not expect to be able to train a complete speech recognizer 
from unlabelled speech and images, it is possible to train models for more constrained tasks, such as cross-modal retrieval~\cite{harwath+etal_nips16,chrupala+etal_acl17}, unsupervised learning of word-like units~\cite{harwath+glass_acl17,harwath+etal_arxiv18}, 
keyword search~\cite{kamper+etal_interspeech17}, and semantic search~\cite{kamper+etal_taslp19}.

Here we explore how unlabelled speech paired with visual context can be used for
{\it semantic query-by-example search} (semantic QbE).
Given a spoken query and a search database of spoken utterances, the task is to find utterances that are semantically relevant to the query.  For example, given a spoken query like ``children'', we would like to retrieve utterances containing the word ``children'' but also utterances {\it about} children, like ``two girls are playing hopscotch.''
This differs from standard QbE, which only seeks exact matches to the query; from keyword spotting and spoken term detection, where the query is written instead of spoken; and from typical semantic search tasks~\cite{lee+etal_taslp15,kamper+etal_taslp19}, which also involve textual~queries.

Our approach to semantic QbE is embedding-based:  We learn an embedding function that maps from segments of speech---queries, search utterances, or sub-segments of search utterances---to fixed-dimensional vectors; we search for semantic matches by {finding the minimum} 
distance between query and search utterance embedding vectors.  In this respect our approach is similar to those in recent embedding-based QbE work~\cite{levin+etal_icassp15,guoguo+etal_icassp15,settle+etal_interspeech17,wang+etal_icassp18}, 
and also some embedding-based spoken term detection work~\cite{audhkhasi2017end}.  The key difference is that our embedding function must be learned in such a way that similar embedding vectors are {\it semantically} rather than {\it phonetically} similar.  For this purpose, training on visually grounded speech data provides the source of semantic information.

Our setting and task are natural to consider for low-resource and even unwritten languages~\cite{adda+etal_sltu16}.  Like much prior work on low-resource methods, in this paper we use English-language data, but we do not use any transcriptions in order 
to simulate a low-resource language setting.

\section{Semantic Q\MakeLowercase{b}E using visual grounding}

\begin{figure}[!b]
    \centering
    \includegraphics[width=0.85\linewidth]{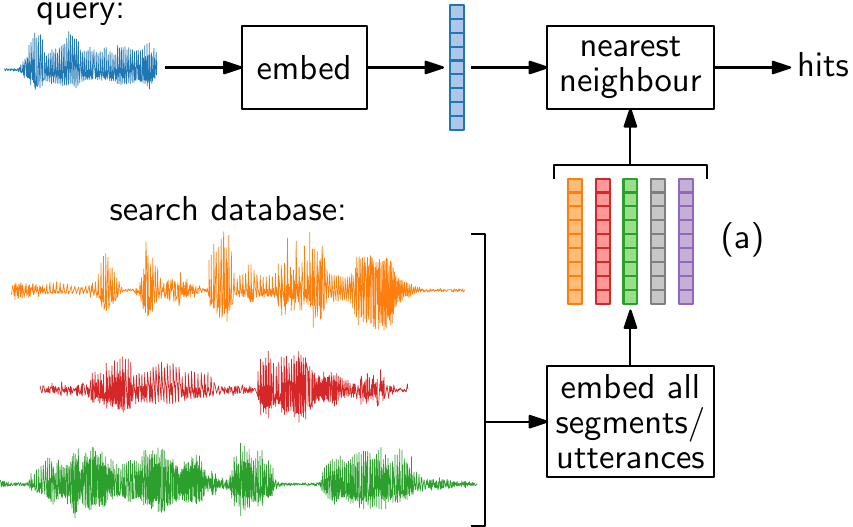}
    \vspace*{-7pt}
    \caption{Embedding-based query-by-example (QbE) search.}
    \label{fig:qbe}
\end{figure}

We perform (semantic) QbE using an embedding-based approach (\S\ref{sec:qbe}), where the acoustic embedding function is obtained through visual grounding~(\S\ref{sec:vision_speech}).
We consider two 
different ways to embed search utterances (\S\ref{sec:fastgrounded} and \S\ref{sec:densegrounded}).

\subsection{Embedding-based QbE}
\label{sec:qbe}

Traditional QbE~\cite{hazen+etal_asru09,zhang+glass_asru09,jansen+vandurme_interspeech12} approaches are based on looking for alignments between the query and spans in the search database, most commonly using dynamic time warping (DTW).  In contrast,
embedding-based QbE~\cite{levin+etal_icassp15,guoguo+etal_icassp15,settle+etal_interspeech17,
  wang+etal_icassp18} 
  relies on an \textit{acoustic embedding}
function, which maps a speech segment (of variable length) to a fixed-dimensional vector.
Ideally, instances of the same word should be mapped to similar vectors while 
unrelated
words (or utterances) should have embeddings that are far apart.
Instead of requiring an alignment between variable-duration segments (as in conventional QbE), queries and search utterances 
(or sub-segments of search utterances) 
are compared directly in this embedding space. 
The overall embedded QbE approach is illustrated in Figure~\ref{fig:qbe}. 

Various acoustic embedding functions have been proposed, with neural models
used in several
studies~\cite{maas+etal_icmlwrl12,levin+etal_asru13,bengio+heigold_interspeech14,kamper+etal_icassp16,chung+etal_interspeech16,settle+livescu_slt16,settle+etal_interspeech17,audhkhasi2017end,chung+glass_interspeech18,holzenberger+etal_interspeech18}. 
Most of these methods, however, require labelled training data.  For example, 
{some recent work uses convolutional or recurrent neural networks learned by optimizing a contrastive loss using a set of known same-word pairs~\cite{kamper+etal_icassp16,settle+livescu_slt16}.} 
Even for studies considering unsupervised acoustic embeddings, true word boundaries are normally used (e.g., in~\cite{levin+etal_asru13,chung+etal_interspeech16}).  
Exceptions include~\cite{audhkhasi2017end,chung+etal_nips18},
which use no annotations.

Here we consider settings where no text labels or any other annotations (such as word boundaries) are available; instead, unlabelled speech is paired with visual context, 
which serves as the sole supervision signal.
We use this visual information to train an acoustic embedding function for use in embedding-based QbE.  
This setting is relevant, e.g., for very low-resource languages or languages without an orthography~\cite{adda+etal_sltu16}.

\vspace*{-5pt}
\subsection{A visually grounded model of speech as the acoustic embedding function}
\label{sec:vision_speech}

\begin{figure}[!t]
    \centering
    \includegraphics[width=0.85\linewidth]{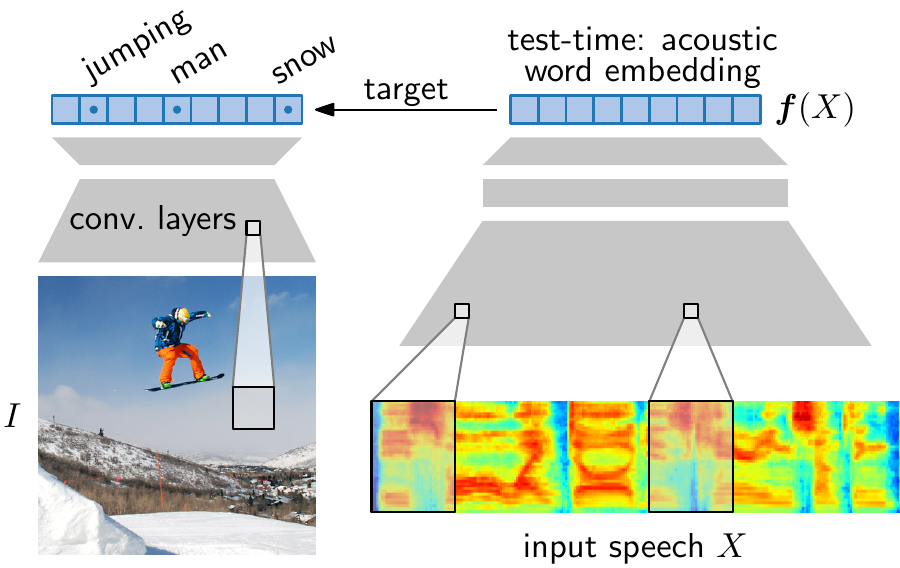}
    \vspace*{-8pt}
    \caption{The acoustic embedding CNN is trained using paired images and unlabelled spoken captions.
    Training targets for the speech network $\vec{f}(X)$ (right) is obtained from the external visual tagger (left).}
    \label{fig:vision_speech_cnn}
    \vspace*{-10pt}
\end{figure}

Given a corpus of images with spoken captions, neither 
having 
textual labels, our goal is to obtain a network that can map an arbitrary length speech segment 
to a fixed-dimensional vector.
Many of the recently proposed vision+speech approaches can be used for this (\S\ref{sec:intro}).  Here we use the method of~\cite{kamper+etal_interspeech17,kamper+etal_taslp19}.
This approach (Figure~\ref{fig:vision_speech_cnn}) 
takes advantage of a separate visual tagger, which predicts relevant text labels for a given input image.
The tagger produces soft keyword labels (posteriors of tags) for each training image in the audio-visual training set.
These are then used as targets for a neural network that maps unlabelled speech to keyword labels.
Without observing any transcriptions, the model can be used to predict which (written) words are present in
a previously
unseen input utterance, acting as a spoken bag-of-words classifier.
This is not possible with most other vision+speech models, which 
{map speech and images into a shared space but do not produce labels.}

Formally, training image $I$ is paired with spoken caption $X = \vec{x}_1, \vec{x}_2, \ldots, \vec{x}_T$, where 
$\vec{x}_t$ is an acoustic feature vector, e.g.\ MFCCs, for frame $t$.
An external vision system (Figure~\ref{fig:vision_speech_cnn}, left) is used to tag $I$ with soft textual labels, giving targets $\hat{\vec{y}}_{\textrm{vis}} \in [0,1]^W$, with $\hat{y}_{\textrm{vis}, w} = P(w | I)$ the estimated probability of word
(tag) $w$ being present in image $I$, and $W$ the number of possible visual tags.
Using 
$\hat{\vec{y}}_{\textrm{vis}}$ as target, we train the speech
model $\vec{f}(X)$ (Figure~\ref{fig:vision_speech_cnn}, right).
This model consists of a CNN over the speech $X$ with a final sigmoidal layer so that $\vec{f}(X) \in [0,1]^W$.
We interpret each dimension of the output as $f_w(X) = P(w | X)$, and train the model using the summed cross-entropy loss (see~\cite{kamper+etal_taslp19}).
Note that $\vec{f}(X)$ is not a distribution over the output
vocabulary since any number of keywords can be present in an
utterance: it is a multi-label classifier where each dimension
$f_w(X)$ can have any value in $[0,1]$. 
Also note that the size-$W$ output vocabulary is implicitly specified by the visual tagger.

After training, $\vec{f}(\cdot)$ can be applied to unseen speech (without any visual input).
For spoken input $X$
of arbitrary duration,
the network output $\vec{f}(X) \in [0,1]^W$ 
is a single $W$-dimensional vector, which we can use as the acoustic
embedding for that input.
We could also use representations from an intermediate layer in the network, 
which could be useful when a specific dimensionality is desired.
We consider both options in \S\ref{sec:exp}.
We can thus use $\vec{f}(\cdot)$ (Figure~\ref{fig:vision_speech_cnn}, right) directly as the embedding function for 
embedding-based QbE (Figure~\ref{fig:qbe}).
A query (Figure~\ref{fig:qbe},
top left) can be fed to the
$\vec{f}(\cdot)$
network and its embedding obtained.
For embedding the search utterances (Figure~\ref{fig:qbe},
bottom left), we consider two options.

\vspace*{-5pt}
\subsection{\system{Fast}: Embed and compare query and search utterances as single vectors} 
\label{sec:fastgrounded}

The first option is to feed an entire
search
utterance to $\vec{f}(\cdot)$, obtaining a single embedding for that utterance.
To determine whether a query occurs in (or is relevant to) an utterance, the query embedding is compared to that single utterance embedding.
Here we use cosine distance
for this comparison.
One disadvantage of this approach is that, even if an instance of the query occurs in the utterance exactly, the utterance embedding will also capture information from all the other words occurring in that utterance.
We use normalization techniques to temper this effect (\S\ref{sec:validation}).
The advantage of this approach is that it is computationally very efficient, which is why we refer to it as \system{Fast}.
For \system{Fast}, there
is thus one embedding for every search utterance at (a) in Figure~\ref{fig:qbe}. The whole-utterance embedding approach has also been used, for example, in embedding-based (written) keyword search~\cite{audhkhasi2017end}. 

\subsection{\system{Dense}: Embed and compare queries to sub-segments within search utterances} 
\label{sec:densegrounded}

\begin{figure}[h]
    \centering
    \includegraphics[width=0.5\linewidth]{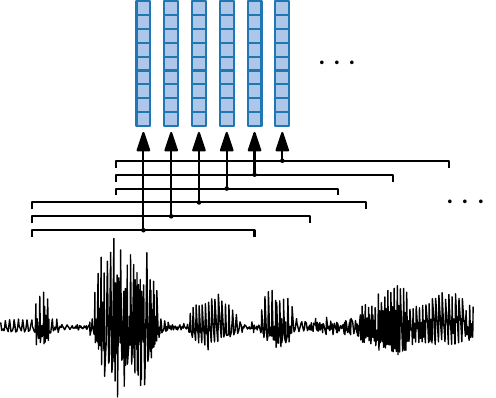}
    \vspace*{-10pt}
    \caption{For \system{Dense}, search utterances are split into overlapping segments which are embedded individually.}
    \label{fig:dense}
    \vspace*{-5pt}
\end{figure}

Instead of obtaining a single embedding for an entire utterance, the \system{Dense} method splits each utterance into overlapping segments from some minimum duration to some maximum duration.
Each segment is then embedded separately using  $\vec{f}(\cdot)$, as illustrated in Figure~\ref{fig:dense}. 
This is similar to the approach used in {some}
previous embedding-based QbE
work~\cite{levin+etal_icassp15,settle+etal_interspeech17}.
\footnote{While~\cite{levin+etal_icassp15,settle+etal_interspeech17} use an
approximate nearest neighbour search procedure, we use exhaustive search here.} 

To determine the relevance of a
search utterance to a query, 
the query embedding is compared to all the embeddings from that utterance.
Specifically, we compare the query to each of the utterance
sub-segment embeddings using cosine distance, and then take the minimum cosine distance as the final score for the relevance of that query to the utterance.
This approach is slower than \system{Fast}, but still much more efficient than performing full alignments between queries and search utterances
using traditional DTW (see~\S\ref{sec:results}).
{\system{Dense} can also predict the location of the segment within an utterance that resulted in a match.  This is not directly possible with \system{Fast}, which scores entire utterances.
However, we do not evaluate localization performance here, and leave this for future work.}
For \system{Dense}, there will therefore be multiple embeddings for each search utterance at (a) in Figure~\ref{fig:qbe}, and this number will depend on the minimum and maximum
segment duration and step size. 

\begin{table*}[!t]
    \mytable
    \caption{Exact and semantic QbE performance on test data.  \system{DTW} performs full alignment.  \system{Grounded} systems use {acoustic
        embeddings} trained only using visual supervision, while \system{Supervised} systems are trained on text labels. \system{Fast} systems represent search utterances as single embeddings, while \system{Dense} systems embed overlapping segments within search utterances.}
    \vspace{5pt}
    \renewcommand{\arraystretch}{1.0}
    \begingroup
    \small
    \begin{tabularx}{1.0\linewidth}{@{}l@{\ \ }lCCCCCCcc@{}}
        \toprule
        & & \multicolumn{3}{c}{Exact QbE (\%)} & \multicolumn{4}{c}{Semantic QbE (\%)} & \multirow{2}{*}{\parbox[t]{1.5cm}{\centering \vspace{0pt} Run-time (min)}} \\
        \cmidrule(lr){3-5} \cmidrule(lr){6-9}
        & Model & \pten & $P@N$ & EER & \pten & $P@N$ & EER & Spearman's $\rho$ &  \\
        \midrule
        \textit{Baselines:}
        & \system{Random} & \leavevmode\hphantom{0}4.5 & \leavevmode\hphantom{0}4.5 & 50\leavevmode\hphantom{.0} & \leavevmode\hphantom{0}9.5 & \leavevmode\hphantom{0}9.1 & 50\leavevmode\hphantom{.0} & \leavevmode\hphantom{0}5.9 & - \\
        & \system{DTW} & 54.6 & 24.9 & 32.1 & 44.3 & 24.3 & 38.7 & 13.7 & 4080\\
        \addlinespace
        \textit{Our systems:} 
        & \system{FastGrounded} & 27.5 & 17.9 & 38.9 & 32.6 & 23.2 & 41.4 & 12.8 & \leavevmode\hphantom{0}$<$ 1 \\
        & \system{DenseGrounded} & 56.0 & 37.3 & 21.7 & 55.5 & 37.3 & 30.0 & 14.9 & \leavevmode\hphantom{0}621\\
        \addlinespace
        \textit{Supervised:}
        & \system{FastSupervised} & 60.7 & 41.3 & 27.2 & 56.6 & 30.9 & 39.8 &  \leavevmode\hphantom{0}8.5 & \leavevmode\hphantom{0}$<$ 1 \\
        & \system{DenseSupervised} & 72.0 & 55.7 & 12.0 & 71.2 & 46.4 & 27.4 & 13.5 & \leavevmode\hphantom{0}568 \\
        \bottomrule
    \end{tabularx}
    \endgroup
    \label{tbl:results}
    \vspace*{-7.5pt}
\end{table*}


\section{Experiments}
\label{sec:exp}

\subsection{Experimental setup and evaluation}

We train our visually grounded acoustic embedding model (\S\ref{sec:vision_speech}) on the corpus of parallel images and spoken captions of~\cite{harwath+glass_asru15}, containing 8000 images with 5 spoken captions each, divided into train, development and test sets.
The audio comprises around 37~hours of active speech in total. 
67 keyword types are selected randomly from transcriptions of the training portion of the corpus.
In the development and test sets, spoken instances of these keywords are extracted as queries, while a disjoint part of each set is used as
a search collection.
There are 
{multiple} queries of the same type, with approximately 2000 spoken queries in total 
being matched to around 4000 search utterances ({in each evaluation set}).
We parametrize speech as 13-dimensional MFCCs with first and second order derivatives. 
Utterances longer than 8~s are~truncated to 8~s. 

The structure and optimization procedure of the visually grounded embedding network (\S\ref{sec:vision_speech}) are the same as in~\cite{kamper+etal_taslp19}: it uses three convolutational layers, one fully connected layer, and a final 1000-unit sigmoid output layer.
We deal with the variable duration of utterances by pooling units over time at the last convolutional layer. 
Each of the $W = \textrm{1000}$ units in the final output corresponds to an image tag from the external visual classifier (also see~\cite{kamper+etal_taslp19}).
{Note that all 67 of the keyword types occurs as one of the tags.}
The output of this speech network is used as embedding function in the \system{Fast} and \system{Dense} {matching variants.}
{To explicitly denote that these systems use a visually grounded embedding method, we denote them as \system{FastGrounded} and \system{DenseGrounded}, respectively.}
For \system{Dense}, a minimum
segment duration of 200~ms and a maximum of 600~ms are used with a step of 30~ms.

As a baseline, we use a simple implementation of a DTW-based QbE system that performs successive alignments: a query is swept over a search utterance (30~ms step size), the DTW alignment cost is calculated over the overlapping segments ({of the same length as the query}), 
and the overall best alignment is taken as the score for how likely that utterance is to contain the query.
More advanced DTW-based QbE systems have been proposed~\cite{park+glass_taslp08,jansen+vandurme_interspeech12} (mainly to improve efficiency), but we restrict ourselves to this exhaustive-search implementation.

We use 3 metrics to quantify how well a QbE system predicts exact query matches~\cite{hazen+etal_asru09,zhang+glass_asru09}: \pten is the average precision (across keywords, in \%) of the $10$ highest-scoring proposals
(utterances); $P@N$ is the average precision of the top $N$ proposals, with $N$ the number of true occurrences of the keyword; and equal error rate (EER) is the average error rate at which false acceptance and false rejection rates are equal.

For 1000 of the test utterances, semantic labels were collected in~\cite{kamper+etal_taslp19} using Amazon Mechanical Turk for the same set of 67 keyword types we use here. 
We use these labels to evaluate semantic QbE performance, where the goal is to retrieve all utterances that are semantically relevant, irrespective of whether an instance of the query occurs exactly in the utterance or not.
Each of the 1000 test utterances was labelled by 5 annotators.
By taking the majority decision, a hard label of whether an utterance is semantically relevant or irrelevant to a query can be assigned. We use these hard labels to calculate semantic \pten, $P@N$ and EER.
We also calculate Spearman's $\rho$, which measures the correlation between a system's ranking and the actual number of annotators that marked a keyword as relevant to an utterance~\cite{agirre+etal_naacl09,hill+etal_cl15}.

\subsection{Results: Exact and semantic QbE}
\vspace*{-3pt}
\label{sec:results}

Table~\ref{tbl:results} shows exact and semantic QbE results. 
A random baseline is included for reference, {which assigns a random relevance score for a search utterance}.
For the \system{Supervised} systems, a speech network was trained on text transcriptions to perform keyword prediction, and embeddings taken from the final output. 
These systems therefore represent the case where perfect text labels are available for training utterances.

\system{FastGrounded} is outperformed by the conventional \system{DTW} QbE approach across all metrics. 
The \system{DenseGrounded} system, however, outperforms \system{DTW} across all metrics.
This also comes with a speed benefit: \system{DenseGrounded} is more than 6 times faster than \system{DTW}. 
(Run-time reported for embedding comparisons
on a single CPU; we parallelized all systems.)
\system{FastGrounded}, which can compare a query and search utterance using a single comparison, is several orders faster than the other approaches, but comes with a cost in performance.

Comparing the exact QbE metrics to the semantic QbE metrics, we see that \system{DTW} and the \system{Supervised} systems all perform worse on semantic QbE.
In contrast, the \system{Grounded} systems perform better on all metrics when moving to semantic QbE.
\system{DenseGrounded} in fact achieves the best overall performance on Spearman's $\rho$, which takes the soft annotator scores into account.
This also aligns with the findings in~\cite{kamper+etal_taslp19}, which considered keyword spotting (where queries are written rather than spoken), and also found that the visually grounded systems aligned better with actual annotator counts.

\vspace*{-3pt}
\subsection{Additional experiments}
\vspace*{-3pt}
\label{sec:validation}

The \system{Dense} systems were tuned on development data to set the maximum and minimum durations and step size of the segments ({although, because of the run-time of these systems, extensive hyper-parameter optimization was not possible}). 

In order to determine what effect lower-dimensional embeddings would have, we also considered a visually grounded embedding network with a penultimate 256-dimensional bottleneck layer; using the bottleneck layer outputs as our acoustic embedding function $\vec{f}(\cdot)$ worsened \pten, $P@N$ and EER by between 5 and 10\% absolute.
Apart from cosine distance
as a measure of embedding distance,
we also considered Euclidean and Kullback-Leibler divergence, but cosine proved best. 

Our original motivation for \system{FastGrounded} was that, if a query contains a keyword of a particular type, the embedding from $\vec{f}(\cdot)$ will have a single dimension with a high probability (since in our case each embedding dimension corresponds to a particular visual tag {and all query types occur as tags}). 
By only considering this specific dimension for all of the search utterance embeddings, a quick retrieval would be possible.
The reasonable performance of \system{FastSupervised} (Table~\ref{tbl:results}) shows that this is in principle possible. 
But we found that when visual grounding is used, embeddings are highly influenced by the prior occurrence of specific visual tags. 
The embedding dimension corresponding to ``man'', for instance, typically has a high score (irrespective of the input), since many training images contain men.
To alleviate this 
effect, we performed mean and variance normalization
on all of the evaluation queries and search utterances using mean and variance estimates from the training embeddings. (We also considered several other normalization methods, but this approach proved most robust.)

\vspace*{-3pt}
\section{Conclusion}
\vspace*{-3pt}

For settings where annotated speech resources are not available, we have shown that query-by-example speech search (QbE) is possible using a model trained on images and unlabelled spoken captions.
Such a model outperforms~a~con\-ventional acoustic alignment-based
(DTW) system, in terms of both exact QbE and semantic QbE, where the goal is to 
{also retrieve non-verbatim matches} related in meaning to the query.
Here we used a specific vision+speech model, but we plan to also investigate how other 
{models (e.g.,~\cite{harwath+etal_nips16})}
can be used to obtain fixed-dimensional acoustic embeddings. 
{There has also been recent work on acoustic-only methods for semantic-acoustic embedding~\cite{chung+glass_interspeech18,chen2018phonetic}, 
which could prove complementary to our approach.}
Finally, we plan to consider how {visual} supervision can be used in truly low-resource languages.\blfootnote{We thank NVIDIA for sponsoring a Titan Xp GPU for this work.
This material is based upon work supported by the Air Force Office of Scientific Research under award number FA9550-18-1-0166.
} 


\newpage
\ninept
\bibliography{icassp2019}

\begin{thebibliography}{10}

\bibitem{synnaeve+etal_slt14}
G.~Synnaeve, T.~Schatz, and E.~Dupoux,
\newblock ``Phonetics embedding learning with side information,''
\newblock in {\em Proc. SLT}, 2014.

\bibitem{palaz+etal_interspeech16}
D.~Palaz, G.~Synnaeve, and R.~Collobert,
\newblock ``Jointly learning to locate and classify words using convolutional
  networks,''
\newblock in {\em Proc. Interspeech}, 2016.

\bibitem{bansal+etal_interspeech18}
S.~Bansal, H.~Kamper, K.~Livescu, A.~Lopez, and S.~Goldwater,
\newblock ``Low-resource speech-to-text translation,''
\newblock in {\em Proc. Interspeech}, 2018.

\bibitem{synnaeve+etal_nipsworkshop14}
G.~Synnaeve, M.~Versteegh, and E.~Dupoux,
\newblock ``Learning words from images and speech,''
\newblock in {\em NIPS Workshop Learn. Semantics}, 2014.

\bibitem{harwath+etal_nips16}
D.~Harwath, A.~Torralba, and J.~R. Glass,
\newblock ``Unsupervised learning of spoken language with visual context,''
\newblock in {\em Proc. NIPS}, 2016.

\bibitem{chrupala+etal_acl17}
G.~Chrupa{\l}a, L.~Gelderloos, and A.~Alishahi,
\newblock ``Representations of language in a model of visually grounded speech
  signal,''
\newblock {\em Proc. ACL}, 2017.

\bibitem{scharenborg+etal_icassp18}
O.~Scharenborg et~al.,
\newblock ``Linguistic unit discovery from multi-modal inputs in unwritten
  languages: Summary of the ``{Speaking Rosetta}'' {JSALT 2017 Workshop},''
\newblock {\em Proc. ICASSP}, 2018.

\bibitem{kamper+roth_sltu18}
H.~Kamper and M.~Roth,
\newblock ``Visually grounded cross-lingual keyword spotting in speech,''
\newblock {\em Proc. SLTU}, 2018.

\bibitem{harwath+glass_acl17}
D.~Harwath and J.~R. Glass,
\newblock ``Learning word-like units from joint audio-visual analysis,''
\newblock in {\em Proc. ACL}, 2017.

\bibitem{harwath+etal_arxiv18}
D.~Harwath et~al.,
\newblock ``Jointly discovering visual objects and spoken words from raw
  sensory input,''
\newblock {\em arXiv preprint arXiv:1804.01452}, 2018.

\bibitem{kamper+etal_interspeech17}
H.~Kamper, S.~Settle, G.~Shakhnarovich, and K.~Livescu,
\newblock ``Visually grounded learning of keyword prediction from untranscribed
  speech,''
\newblock in {\em Proc. Interspeech}, 2017.

\bibitem{kamper+etal_taslp19}
H.~Kamper, G.~Shakhnarovich, and K.~Livescu,
\newblock ``Semantic speech retrieval with a visually grounded model of
  untranscribed speech,''
\newblock {\em IEEE Trans. Audio, Speech, Language Process.}, vol. 27, no. 1,
  pp. 89--98, 2019.

\bibitem{lee+etal_taslp15}
L.-s. Lee, J.~Glass, H.-y. Lee, and C.-a. Chan,
\newblock ``Spoken content retrieval---beyond cascading speech recognition with
  text retrieval,''
\newblock {\em IEEE Trans. Audio, Speech, Language Process.}, vol. 23, no. 9,
  pp. 1389--1420, 2015.

\bibitem{levin+etal_icassp15}
K.~Levin, A.~Jansen, and B.~Van~Durme,
\newblock ``Segmental acoustic indexing for zero resource keyword search,''
\newblock in {\em Proc. ICASSP}, 2015.

\bibitem{guoguo+etal_icassp15}
G.~Chen, C.~Parada, and T.~N. Sainath,
\newblock ``Query-by-example keyword spotting using long short-term memory
  networks,''
\newblock in {\em Proc. ICASSP}, 2015.

\bibitem{settle+etal_interspeech17}
S.~Settle, K.~Levin, H.~Kamper, and K.~Livescu,
\newblock ``Query-by-example search with discriminative neural acoustic word
  embeddings,''
\newblock in {\em Proc. Interspeech}, 2017.

\bibitem{wang+etal_icassp18}
Y.-H. Wang, H.-y. Lee, and L.-s. Lee,
\newblock ``Segmental audio word2vec: Representing utterances as sequences of
  vectors with applications in spoken term detection,''
\newblock in {\em Proc. ICASSP}, 2018.

\bibitem{audhkhasi2017end}
K.~Audhkhasi, A.~Rosenberg, A.~Sethy, B.~Ramabhadran, and B.~Kingsbury,
\newblock ``End-to-end asr-free keyword search from speech,''
\newblock {\em IEEE J. Sel. Topics Signal Process.}, vol. 11, no. 8, pp.
  1351--1359, 2017.

\bibitem{adda+etal_sltu16}
G.~Adda et~al.,
\newblock ``Breaking the unwritten language barrier: The {BULB} project,''
\newblock {\em Proc. SLTU}, 2016.

\bibitem{hazen+etal_asru09}
T.~J. Hazen, W.~Shen, and C.~White,
\newblock ``Query-by-example spoken term detection using phonetic posteriorgram
  templates,''
\newblock in {\em Proc. ASRU}, 2009.

\bibitem{zhang+glass_asru09}
Y.~Zhang and J.~R. Glass,
\newblock ``Unsupervised spoken keyword spotting via segmental {DTW} on
  {G}aussian posteriorgrams,''
\newblock in {\em Proc. ASRU}, 2009.

\bibitem{jansen+vandurme_interspeech12}
A.~Jansen and B.~Van~Durme,
\newblock ``Indexing raw acoustic features for scalable zero resource search,''
\newblock in {\em Proc. Interspeech}, 2012.

\bibitem{maas+etal_icmlwrl12}
A.~L. Maas, S.~D. Miller, T.~M. O'Neil, A.~Y. Ng, and P.~Nguyen,
\newblock ``Word-level acoustic modeling with convolutional vector
  regression,''
\newblock in {\em Proc. ICML Workshop Representation Learn.}, 2012.

\bibitem{levin+etal_asru13}
K.~Levin, K.~Henry, A.~Jansen, and K.~Livescu,
\newblock ``Fixed-dimensional acoustic embeddings of variable-length segments
  in low-resource settings,''
\newblock in {\em Proc. ASRU}, 2013.

\bibitem{bengio+heigold_interspeech14}
S.~Bengio and G.~Heigold,
\newblock ``Word embeddings for speech recognition,''
\newblock in {\em Proc. Interspeech}, 2014.

\bibitem{kamper+etal_icassp16}
H.~Kamper, W.~Wang, and K.~Livescu,
\newblock ``Deep convolutional acoustic word embeddings using word-pair side
  information,''
\newblock in {\em Proc. ICASSP}, 2016.

\bibitem{chung+etal_interspeech16}
Y.-A. Chung, C.-C. Wu, C.-H. Shen, and H.-Y. Lee,
\newblock ``Unsupervised learning of audio segment representations using
  sequence-to-sequence recurrent neural networks,''
\newblock {\em in Proc. Interspeech}, 2016.

\bibitem{settle+livescu_slt16}
S.~Settle and K.~Livescu,
\newblock ``Discriminative acoustic word embeddings: Recurrent neural
  network-based approaches,''
\newblock in {\em Proc. SLT}, 2016.

\bibitem{chung+glass_interspeech18}
Y.-A. Chung and J.~R. Glass,
\newblock ``Speech2vec: A sequence-to-sequence framework for learning word
  embeddings from speech,''
\newblock in {\em Proc. Interspeech}, 2018.

\bibitem{holzenberger+etal_interspeech18}
N.~Holzenberger, M.~Du, J.~Karadayi, R.~Riad, and E.~Dupoux,
\newblock ``Learning word embeddings: Unsupervised methods for fixed-size
  representations of variable-length speech segments,''
\newblock {\em in Proc. Interspeech}, 2018.

\bibitem{chung+etal_nips18}
Y.-A. Chung, W.-H. Weng, S.~Tong, and J.~R. Glass,
\newblock ``Unsupervised cross-modal alignment of speech and text embedding
  spaces,''
\newblock in {\em Proc. NIPS}, 2018.

\bibitem{harwath+glass_asru15}
D.~Harwath and J.~R. Glass,
\newblock ``Deep multimodal semantic embeddings for speech and images,''
\newblock in {\em Proc. ASRU}, 2015.

\bibitem{park+glass_taslp08}
A.~S. Park and J.~R. Glass,
\newblock ``Unsupervised pattern discovery in speech,''
\newblock {\em IEEE Trans. Audio, Speech, Language Process.}, vol. 16, no. 1,
  pp. 186--197, 2008.

\bibitem{agirre+etal_naacl09}
E.~Agirre et~al.,
\newblock ``A study on similarity and relatedness using distributional and
  wordnet-based approaches,''
\newblock in {\em Proc. HLT-NAACL}, 2009.

\bibitem{hill+etal_cl15}
F.~Hill, R.~Reichart, and A.~Korhonen,
\newblock ``Sim{L}ex-999: Evaluating semantic models with (genuine) similarity
  estimation,''
\newblock {\em Comput. Linguist.}, vol. 41, no. 4, 2015.

\bibitem{chen2018phonetic}
Y.-C. Chen, S.-F. Huang, C.-H. Shen, H.-y. Lee, and L.-s. Lee,
\newblock ``Phonetic-and-semantic embedding of spoken words with applications
  in spoken content retrieval,''
\newblock {\em arXiv preprint arXiv:1807.08089}, 2018.

\end{thebibliography}

\end{document}